\documentclass[letterpaper,conference]{IEEEtran}  

\IEEEoverridecommandlockouts                              

\usepackage{graphics} 
\usepackage{hyperref}
\usepackage{amsmath, amsfonts, amssymb, amscd, bbm} 
\usepackage[usenames,dvipsnames]{color} 
\usepackage[capitalize]{cleveref}
\usepackage{longtable}
\usepackage{pdfpages}
\usepackage{pgfplots}
\pgfdeclarelayer{background}
\pgfdeclarelayer{foreground}
\pgfsetlayers{background,main,foreground}
\usepackage{tikz}
\usepackage{aircraftshapes}
\usetikzlibrary{arrows}
\usetikzlibrary{positioning}
\usepackage{nomencl}
\usepackage{subcaption}
\makenomenclature
\usepackage[per-mode=symbol,mode=text,group-separator={,},group-four-digits=true,load=accepted,list-final-separator={, and }]{siunitx}
\usepackage{adjustbox}
\usepackage{multirow}
\usepackage{float}
\floatstyle{plaintop}
\restylefloat{table}
\usepackage{footnote}
\usepackage{threeparttable}
\usepackage{arydshln}

\usepackage{booktabs}

\usepackage[labelsep=period]{caption}

\newunit{\hour}{hours}
\newunit{\feet}{ft}
\newunit{\nauticalmile}{NM}
\newunit{\gravity}{g}
\newunit{\mph}{mph}
\newunit{\knots}{knots}

\pgfplotsset{compat=newest}
\usepackage[backend=bibtex,style=ieee]{biblatex}
\AtEveryBibitem{
\ifentrytype{inproceedings}{
    \clearlist{address}
    \clearlist{publisher}
    \clearname{editor}
    \clearlist{organization}
    \clearfield{url}  
    \clearfield{pages}  
    \clearlist{location}
 }{}
 }
 \AtEveryBibitem{
 \clearfield{doi}
 }

\DeclareMathOperator*{\argmax}{arg\,max}

\hypersetup{
    colorlinks=true,
    hypertexnames=false,
    plainpages = false,
    breaklinks=true,
    linkcolor = black,  
    anchorcolor = black,
    citecolor = black,
    filecolor = black,
    menucolor = black,
    urlcolor = black,
    pdfauthor={Sheng Li, Maxim Egorov, Mykel J. Kochenderfer},
    pdftitle={Optimizing Collision Avoidance in Dense Airspace using DeepReinforcement Learning},
    pdfkeywords={Separation; Autonomous systems and operations; Air Traffic Control; Markov Decision Process.}
}

\usepackage[ruled]{algorithm2e}

\title{
\small{Thirteenth USA/Europe Air Traffic Management Research and Development Seminar (ATM2019)}\\
\LARGE \bf Optimizing Collision Avoidance in Dense Airspace \\ using Deep Reinforcement Learning
}

\author{\IEEEauthorblockN{Sheng Li}
\IEEEauthorblockA{Aeronautics and Astronautics\\
Stanford University\\
Stanford, California, USA\\
lisheng@stanford.edu}
\and
\IEEEauthorblockN{Maxim Egorov}
\IEEEauthorblockA{Airbus UTM\\
San Francisco, California, USA\\
maxim.egorov@airbus-sv.com}
\and
\IEEEauthorblockN{Mykel J. Kochenderfer}
\IEEEauthorblockA{Aeronautics and Astronautics\\
Stanford University\\
Stanford, California, USA\\
mykel@stanford.edu}}

\begin{document}

\maketitle
\thispagestyle{empty}
\pagestyle{empty}

\begin{abstract}

New methodologies will be needed to ensure the airspace remains safe and efficient as traffic densities rise to accommodate new unmanned operations. 
This paper explores how unmanned free-flight traffic may operate in dense airspace.  
We develop and analyze autonomous collision avoidance systems for aircraft operating in dense airspace where traditional collision avoidance systems fail.
We propose a metric for quantifying the decision burden on a collision avoidance system as well as a metric for measuring the impact of the collision avoidance system on airspace.
We use deep reinforcement learning to compute corrections for an existing collision avoidance approach to account for dense airspace. 
The results show that a corrected collision avoidance system can operate more efficiently than traditional methods in dense airspace while maintaining high levels of safety.

\end{abstract}


\section{Introduction}


Recent technological advances have enabled a number of new applications for unmanned aircraft ranging from aerial cargo delivery to autonomous vertical take-off and landing (VTOL) passenger aircraft. 
It is estimated that by the year 2035, the number of package delivery aircraft in the sky will increase by one to two orders of magnitude~\cite{jenkins2017forecast}, while the number of passenger carrying VTOL operations is expected to increase at a similar pace~\cite{karthik2018blueprint}.
This increase will lead to hundreds or even thousands of aircraft occupying 
relatively small volumes of airspace, and will require new methodologies to ensure safe and efficient operations. 

It is unclear how traditional air traffic management (ATM) approaches for maintaining safety and efficiency in the airspace perform in the context of high-volume unmanned traffic. 
There has been tremendous interest in on-board collision avoidance systems (CAS), both in the context of manned commercial aviation~\cite{mcfadyen2013aircraft,kochenderfer2012next} and in the context of unmanned aircraft~\cite{thipphavong2017ensuring,mueller2016multi}. 
For example, the Traffic-alert and Collision Avoidance System (TCAS) was designed for manned aviation and can accommodate densities of up to 0.3~aircraft/nmi$^2$~\cite{williamson1989development}.
Its successor, the next-generation airborne collision avoidance system (ACAS X) formulates the CAS problem as a partially observable Markov decision process (POMDP) and is able to operate in even denser airspace~\cite{kochenderfer2011robust,kochenderfer2012next}.
While ACAS X has been extended to unmanned aircraft~\cite{manfredi2016introduction} and resolving conflicts with multiple threats~\cite{chryssanthacopoulos2012decomposition}, its performance in ultra-dense airspace has not been deeply studied, with evaluations primarily focused on the much more common pairwise aircraft encounters~\cite{julian2016policy,davies2018comparative}. 

Collision avoidance has been studied in fields outside of aviation with applications ranging from robotics~\cite{van2011reciprocal} to autonomous vehicles~\cite{mukhtar2015vehicle}. 
When communication networks exist, the problem can be solved using centralized path optimization~\cite{chen2015decoupled, tang2015mixed}. 
A number of decentralized approaches have also been developed to solve sequential multi-agent decision problems using deep reinforcement learning (DRL)~\cite{foerster2017stabilising,foerster2018counterfactual,gupta2017cooperative}, which can scale to large observation spaces and many agents. 
DRL has been extended to collision avoidance through approaches that learn interaction dynamics~\cite{chen2017decentralized}, explicitly model dynamic uncertainty~\cite{kahn2017uncertainty}, and learn policies end-to-end~\cite{long2018towards}.
However, the performance of collision avoidance strategies typically degrades when the number of agents increases due to an exponential growth 
in the state space. 
Designing CAS policies for high airspace densities will require a new set of approaches, and this paper aims to explore one of them.


We formulate collision avoidance as a stochastic problem in the form of a multi-agent Markov decision process (MMDP) similar to~\cite{ong2016markov} with a focus on resolutions in the horizontal plane.
On top of decomposing the problem into pairwise encounters, we apply a DRL based approach to improve the collision avoidance in dense airspace.
We combine the decentralized training approach that has shown to scale in multi-agent systems~\cite{gupta2017cooperative,long2018towards} with a deep correction factor~\cite{bouton2018utility} to explicitly capture the properties of a multi-agent system and the requirements for collision avoidance. 
The contributions of this work are as follows: 
(1) an approach that adds corrections learned through DRL to an existing policy for further improving collision avoidance in dense airspace,
(2) an analysis of how collision avoidance systems impact operations in the dense airspace,
(3) recommendations for how to approach the dense airspace problem from the perspective of collision avoidance.

\section{Problem Formulation}
This section introduces the mathematical framework for collision avoidance using the Markov decision process (MDP)~\cite{bellman2015applied}. 

\subsection{Markov Decision Process}
An MDP is formally defined by the tuple ($\mathcal{S}$, $\mathcal{A}$, $T$, $R$, $\gamma$), where $\mathcal{S}$ is the state space, $\mathcal{A}$ is the action space, $T$ is the state transition function, $R$ is the reward function, and $\gamma$ is the discount factor. 
In an MDP, an agent takes action $a_t \in \mathcal{A}$ at time $t$ based on the state $s_t \in \mathcal{S}$, and receives a reward $r_t = R(s_t, a_t)$. 
At time $t + 1$, the state transits from $s_t$ to $s_{t+1}$ with a probability $\Pr(s_{t+1} \mid s_t, a_t) = T (s_{t+1}, s_t, a_t)$. 
The objective of the agent is to maximize the accumulated expected discounted reward $\sum_{t=0}^{\infty}\gamma^t r_t$.

A solution to an MDP is a policy $\pi : \mathcal{S} \to \mathcal{A}$ that defines what action to execute at a given state. 
An optimal policy $\pi^*$ of an MDP can be represented by a state-action value function $Q^*(s,a)$ that satisfies the Bellman equation~\cite{kochenderfer2015decision}:
\begin{equation}\label{eq:bellman_transition}
    Q^*(s,a) = R(s,a) + \gamma \sum_{s'}T(s',s,a)\max_{a'}Q^*(s',a')\text{,}
\end{equation}
where $s$ is the current state and $s'$ is a state reachable at the next time step by taking action $a$. 
In this work we use sigma-point sampling~\cite{bertuccelli2008robust} and a generative model to formulate the transition function, which allows us to re-write the Bellman equation in a more general form:
\begin{equation}\label{eq:bellman_expectation}
    Q^*(s,a) = \mathbb{E}_{s'}\left[R(s,a) + \gamma \max_{a'}Q^*(s',a')\right]\text{,}
\end{equation}
which represents the expected discounted reward for the next state $s'$. 
With $Q^*$, the corresponding optimal policy can be written as $\pi^*(s) = \arg \max_a Q^*(s,a)$.
While the optimal utility is given by $U^*(s) = \max_a Q^*(s,a)$.

\subsection{Dynamics and Sensor Measurements}\label{sec:sensor}

In this work we focus on co-altitude, horizontal encounters.
The dynamics of the aircraft are described by its position coordinates $(x,y)$, speed $v$, heading angle $\phi$ and turn rate $\dot{\phi}$, and are updated by
\begin{align}
\label{eqn:dyn_update}
\begin{split}
    \phi &\leftarrow \phi + \dot{\phi}\Delta t\text{,} \\
    x &\leftarrow x + v \cos{\phi} \Delta t\text{,} \\
    y &\leftarrow y + v \sin{\phi} \Delta t.
\end{split}
\end{align}

\begin{figure}
    \centering
    \includegraphics[width=0.4\textwidth]{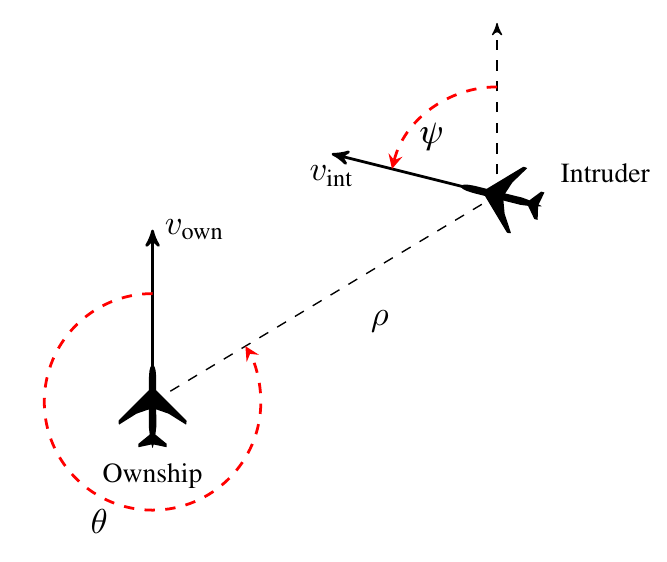}
    \caption{Sensor measurements for a co-altitude pairwise encounter~\cite{julian2016policy}.}
    \label{fig:sensor}
\end{figure}

The sensor model in our aircraft can be described by the following variables:
\begin{enumerate}
    \item $\rho$: Distance from the ownship to the intruder.
    \item $\theta$: Angle to the intruder relative to ownship heading direction.
    \item $\psi$: Heading angle of the intruder relative to the heading direction of the ownship.
    \item $v_\text{own}$: Speed of the ownship.
    \item $v_\text{int}$: Speed of the intruder.
\end{enumerate}
An example encounter that includes the sensor measurements is illustrated in Fig.~\ref{fig:sensor}.

\subsection{Action Space}

The collision avoidance policy can issue the following advisories to resolve conflicts: \{clear of conflict (COC), weak left (WL), weak right (WR), strong left (SL), strong right (SR), MAINTAIN\}. 
These advisories can be transformed into turn rates that control the aircraft in the following way:
\begin{align*}
    \mathcal{A} = \{&\text{COC} \rightarrow \text{free to fly towards the destination}, \\
    &\text{WL} \rightarrow + 5^\circ/\text{s}, \text{WR} \rightarrow -5^\circ/\text{s}, \\
    &\text{SL} \rightarrow + 10^\circ/\text{s}, \text{SR} \rightarrow -10^\circ/\text{s}, \text{MAINTAIN} \rightarrow 0 \}.
\end{align*}
The above discretization of turn rates was chosen to allow realistic control over the decision period considered in this paper. 

\section{CAS for Conventional Traffic Density}

We use dynamic programming to compute pairwise conflict resolution policies similar to the approach for ACAS X~\cite{kochenderfer2012next}. 
Using the pairwise policies, we apply utility decomposition to approximate the optimal policies for multi-threat conflict resolution.

\subsection{Pairwise Conflict Resolution}\label{sec:pairwise}
A co-altitude pairwise encounter is illustrated in Fig.~\ref{fig:sensor}, i.e. there is only one intruder within the sensing range of the ownship. 
We define the state space, state transition and the reward function for this pairwise encounter below.

\subsubsection{State Space}
The state space for a pairwise encounter is composed of a discrete set of locations, headings, and speeds of the intruder relative to the ownship. 
A single state $s$ represented by the vector
$[\rho, \theta, \psi, v_\text{own}, v_\text{int}]$, where each dimension of the state is discretized into finite grids.

\subsubsection{State Transition} \label{sec:transition_sampling}
The state transition function comes from updating the dynamics of the ownship-intruder pair.
We use sigma-point sampling to add noise to speed $v$ and turn rate $\dot{\phi}$ in the dynamics model~\cite{ong2015short}.

\subsubsection{Reward Function}

The objective of the policy is to resolve a conflict while maintaining safety and efficiency. 
To enforce this trade-off, we discourage aircraft being in close proximity to each other, and penalize large and frequent alerts.
The reward function is 
\begin{equation}
    \begin{aligned}
        R(s, a) = & -w_\rho \exp\left( \frac{-(\rho(s) - \rho_\text{NMAC})}{\rho_\text{NMAC}} \right) - w_a \text{turnrate}(a)^2  \\
        & - w_\text{NMAC} \mathbbm{1}\{\rho(s) \leqslant \rho_\text{NMAC}\}  - w_\text{conflict} \mathbbm{1}\{a \neq \text{COC}\}\text{,}
    \end{aligned}
\end{equation}
where $w_\rho$ penalizes close distance between the ownship and the intruder ($\rho_\text{NMAC}$ is a predefined threshold for NMAC), $w_a$ penalizes large magnitude of turn rate (the turn rate of COC is defined to be zero), $w_\text{NMAC}$ penalizes the occurrence of NMAC, $w_\text{conflict}$ penalizes alerts.

\subsubsection{Value Iteration}
As an MDP, the pairwise conflict problem can be solved using a dynamic programming approach known as value iteration~\cite{kochenderfer2015decision}. 
The idea is to iteratively optimize the state-action value function $Q(s,a)$ for all $s$ and $a$ using the update
\begin{equation}
    Q_{k+1}(s,a) \leftarrow R(s,a) + \gamma\sum_{s'}T(s',s,a)\max_{a'} Q_k(s',a').
\end{equation}
The result of value iteration is an optimal state-action value function $Q^*(s,a)$.

 

\subsubsection{Policy}
In the context of pairwise conflict resolution, $Q^*(s,a)$ acts as a numeric table for the ownship, which takes in the state and returns the evaluation on each action.
We can extract an optimal policy $\pi^*(s)$ by using the lookup $\pi^*(s) = \argmax_a Q^*(s,a)$.

\subsection{Multi-threat Conflict Resolution}\label{sec:multi}
For conflicts with more than one intruder, the globally optimal solution would involve solving a single multi-agent MDP that takes all the intruders into consideration. 
However, this approach would be hard to scale since the dimension of the state space would grow exponentially with the number of intruders.
Instead, we can combine simple sub-problems to approximate the complete multi-agent solution in a more efficient way. 

\begin{figure}
\centering
\small
\begin{tikzpicture}[
>=stealth',
Qi/.style={rectangle, draw=black, fill=white, minimum width=14pt, minimum height=12pt},
Qlo/.style={rectangle, draw=black!70, thick, dashed, fill=black!10, minimum width=90pt, minimum height=80pt},
f/.style={rectangle, draw=black, fill=white, minimum width=20pt, minimum height=14pt}
]
\node[]    ()            at (0, 1.5)   {Global State $s$};
\node[]    (globalState) at (0, 0)     {$\Bigg[\vdots\Bigg]$};
\node[Qlo] ()            at (3.1, 0)   {};
\node[Qi]  (Q1)          at (2, 1)     {$Q_1^*$};
\node[Qi]  (Q2)          at (2, 0.3)   {$Q_2^*$};
\node[]    ()            at (2, -0.25) {$\vdots$};
\node[Qi]  (Qn)          at (2, -1)    {$Q_n^*$};
\node[f]   (fusion)      at (3.6, 0)   {\begin{tabular}{c}Fusion $f$ \\ min/sum \end{tabular}};
\node[]    ()            at (6.1, 1.3) {\begin{tabular}{c}Approximated \\ $Q^*$ \end{tabular}};
\node[]    (globalQ)     at (6.1, 0)   {$\Bigg[\vdots\Bigg]$};

\draw[->] (globalState) to node[above] {$s_1$} (Q1.west);
\draw[->] (globalState) to node[above] {$s_2$} (Q2.west);
\draw[->] (globalState) to node[above] {$s_n$} (Qn.west);
\draw[->] (Q1.east) to node[above] {} (fusion.west);
\draw[->] (Q2.east) to node[above] {} (fusion.west);
\draw[->] (Qn.east) to node[above] {} (fusion.west);
\draw[->] (fusion) to (globalQ);

\end{tikzpicture}
\caption{Utility decomposition~\cite{bouton2018utility}.}
\label{fig:decomposition}
\end{figure}
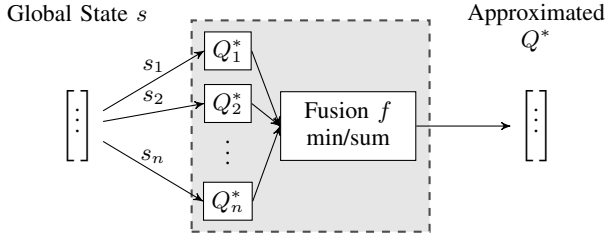
We use utility decomposition~\cite{ong2015short, chryssanthacopoulos2012decomposition} to split a non-cooperative multi-threat conflict resolution problem into pairwise conflict resolution sub-problems.
Let $Q_i^*(s_i,a)$ denote the optimal state-action value function of sub-problem $i$. 
We assume that the state of the full problem $s$ contains the information needed by the state of each sub-problem $s_i$. 
The optimal state-action value function for the full problem $Q^*(s,a)$ can then be approximated by
\begin{equation}
    Q^*(s,a) \approx f(Q_1^*(s_1,a), Q_2^*(s_2,a), \dots, Q_n^*(s_n,a))\text{,}
\end{equation}
where function $f$ performs utility fusion~\cite{bouton2018utility}. 
We consider two approaches to utility fusion.
A summation over state-action values which can be written as $Q^*(s,a) \approx \sum_{i}Q_i^*(s_i,a)$.
Another approach is to considers the intruder with the lowest state-action value, i.e. the intruder with the highest threat level $Q^*(s,a) \approx \min_i Q_i^*(s_i,a)$.
Taking the minimum value is considered a risk averse strategy~\cite{bouton2018utility}. 
Fig.~\ref{fig:decomposition} illustrates the mechanism of utility decomposition and approximation through utility fusion.

The policy of multi-threat conflict resolution can be extracted from the approximated state-action value function $Q^*(s,a)$ by choosing the action with maximum value. 
When this approach combined with the summation and the minimization approaches above, we refer to them as max-sum and max-min respectively.
Prior work has shown that the max-min is superior over max-sum in terms of safety performance\cite{ong2015short,bouton2018utility}. 
We adopt max-min as our decomposition method.

\section{Collision Avoidance in Dense Airspace}
In this section, we outline how collision avoidance can be further improved for operations in dense airspace over existing utility decomposition methods through policy correction.

\subsection{Policy Correction}
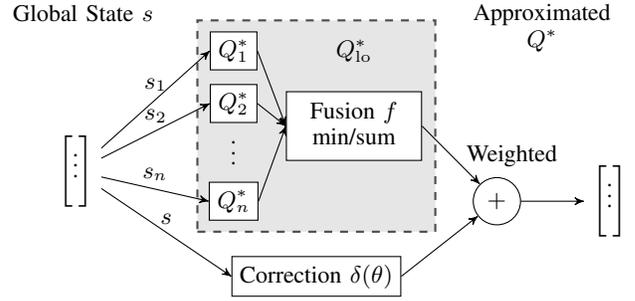
\begin{figure}
\centering
\small
\begin{tikzpicture}[
>=stealth',
Qi/.style={rectangle, draw=black, fill=white, minimum width=14pt, minimum height=12pt},
Qlo/.style={rectangle, draw=black!70, thick, dashed, fill=black!10, minimum width=90pt, minimum height=80pt},
f/.style={rectangle, draw=black, fill=white, minimum width=20pt, minimum height=14pt},
oper/.style={circle, draw=black, fill=white, minimum size=13pt}
]
\node[]    ()            at (-0.1, 1.5)   {Global State $s$};
\node[]    (globalState) at (-0.2, -0.6)  {$\Bigg[\vdots\Bigg]$};
\node[Qlo] ()            at (3, 0)   {};
\node[Qi]  (Q1)          at (1.9, 1)     {$Q_1^*$};
\node[Qi]  (Q2)          at (1.9, 0.3)   {$Q_2^*$};
\node[]    ()            at (1.9, -0.25) {$\vdots$};
\node[Qi]  (Qn)          at (1.9, -1)    {$Q_n^*$};
\node[f]   (fusion)      at (3.5, 0)   {\begin{tabular}{c}Fusion $f$\\ min/sum \end{tabular}};
\node[]    (Qlo)         at (3.5, 1)   {$Q_\mathrm{lo}^*$};
\node[oper](add)         at (5.4, -1) {$+$};
\node[]    ()            at (6.0, 1.3) {\begin{tabular}{c}Approximated \\ $Q^*$ \end{tabular}};
\node[]    ()            at (5.6, -0.4) {Weighted};
\node[]    (globalQ)     at (6.9, -1)   {$\Bigg[\vdots\Bigg]$};
\node[Qi] (correction) at (3, -2.0) {Correction $\delta(\theta)$};

\draw[->] (globalState) to node[above] {$s_1$} (Q1.west);
\draw[->] (globalState) to node[above] {$s_2$} (Q2.west);
\draw[->] (globalState) to node[above] {$s_n$} (Qn.west);
\draw[->] (Q1.east) to (fusion.west);
\draw[->] (Q2.east) to (fusion.west);
\draw[->] (Qn.east) to (fusion.west);
\draw[->] (fusion.east) to (add);
\draw[->] (add) to (globalQ);
\draw[->] (globalState) to node[above] {$s$} (correction.west);
\draw[->] (correction.east) to (add);

\end{tikzpicture}
\caption{Utility decomposition with correction~\cite{bouton2018utility}.}
\label{fig:correction}
\end{figure}

The formulation of policy correction can be derived from multi-fidelity optimization~\cite{bouton2018utility}. 
When a high-fidelity model ($f_\text{hi}$) is too expensive to evaluate, a surrogate model can be used.
The surrogate combines a simpler low-fidelity model ($f_\text{lo}$) and an additive parametric correction term ($\delta$) to approximate $f_\text{hi}$ as $f_\text{hi} \approx f_\text{lo} + \delta$.

In the context of multi-threat conflict resolution, the global optimal solution to the full problem $Q^*(s,a)$ is unfeasible to solve for.
However, we can get a low-fidelity solution $Q^*_\text{lo}(s,a)$ using utility decomposition.
We then add a parameterized correction term to approximate $Q^*(s,a)$ by
\begin{equation}
    Q^*(s,a) \approx (1-w_\text{c})Q^*_\text{lo}(s,a) + w_\text{c} \delta(s,a;\theta)\text{,}
\end{equation}
where $\delta(s,a;\theta)$ is the correction term parameterized by $\theta$, and $w_\text{c}$ is the weight placed on the correction.
\cref{fig:correction} shows the mechanism of adding correction to utility decomposition.

\subsection{Deep Correction Network}

We use the deep Q-network (DQN)~\cite{mnih2015human} to learn the parameters $\theta$ for the correction term $\delta(s,a;\theta)$. 
DQN uses a neural network to approximate the state-action value function of an MDP. 
It can be expressed as $Q(s,a;\theta)$, where $\theta$ represents the weights of the neural network. 
The parameters of a DQN policy can be computed by minimizing the cost function $J$ based on the temporal difference:
\begin{equation}
    J(\theta) = \mathbb{E}_{s'}\left[ \left(r + \gamma \max_{a'}Q(s',a';\theta_-) - Q(s,a;\theta) \right)^2 \right]\text{,}
\end{equation}
where $r=R(s,a)$, and $\theta_-$ defines a fixed target network to be updated periodically with new parameters $\theta$.
The loss is minimized using experience samples $(s,a,r,s')$ that are collected during simulation.
The update rule for $\theta$ is
\begin{equation}\label{eq:update}
    \theta \leftarrow \theta - \alpha\left[r + \gamma \max_{a'}Q(s',a';\theta_-) - Q(s,a;\theta)\right]\nabla_\theta Q(s,a;\theta)\text{,}
\end{equation}
where $\alpha$ is a configurable hyperparameter known as the learning rate. 

By representing the correction as a neural network, we can learn it directly using DQN in a process known as deep correction.
We use the utility decomposition policy as a fixed low-fidelity approximation for the optimal multi-threat policy.
With some modification, the update rule becomes:
\begin{equation}
    \begin{aligned}
        \theta \leftarrow \theta - \alpha &\Big[r + \gamma \max_{a'}\big((1-w_\text{c}) Q_\text{lo}(s',a') + w_\text{c} \delta(s',a';\theta_-)\big)\\
        &-\big((1-w_\text{c}) Q_\text{lo}(s,a) + w_\text{c} \delta(s,a;\theta)\big) \Big]\nabla_\theta \delta(s,a;\theta).
    \end{aligned}
\end{equation}
We use a simulator to train the correction network using the modified update rule.


\subsection{State Space}\label{sec:full_state}
When training the deep correction network, we include additional information in the observation as input into the policy. 
We define the two approaches below. 

\subsubsection{Closest Intruders in Sectors} 
\begin{figure}[t]
    \centering
    \small
    \begin{tikzpicture}[]
        \draw[black,thick,dashed] (0, 0) circle (2);
        \draw[blue!50,thick,dashed] (1.734, 1) -- (-1.734, -1);
        \draw[blue!50,thick,dashed] (-1, 1.734) -- (1, -1.734);
        \node[aircraft top,fill=black,minimum width=0.4cm, rotate=30] (ownship) at (0,0) {};
        \node[aircraft top,fill=red,minimum width=0.4cm, rotate=225] (int11)   at (0.8,0.8) {};
        \node[aircraft top,fill=green,minimum width=0.4cm, rotate=200] (int12)   at (0.8,1.6) {};
        \node[aircraft top,fill=green,minimum width=0.4cm, rotate=275] (int13)   at (-0.24,1.76) {};
        \node[aircraft top,fill=red,minimum width=0.4cm, rotate=0] (int21)   at (-1.6, 0.16) {};
        \node[aircraft top,fill=red,minimum width=0.4cm, rotate=15] (int41)   at (1.2,-0.24) {};
        \node[aircraft top,fill=green,minimum width=0.4cm, rotate=155] (int42)   at (1.28,-0.8) {};
        \node[aircraft top,fill=red,minimum width=0.4cm, rotate=15] (int31)   at (-0.8,-0.8) {};
        \node[] (1)   at (0.129,0.483) {1};
        \node[] (2)   at (-0.483,0.129) {2};
        \node[] (3)   at (-0.129,-0.483) {3};
        \node[] (4)   at (0.483,-0.129) {4};
        \node[aircraft top,fill=black,minimum width=0.4cm] () at (3,2.2) {};
        \node[] () at (4,2.2) {Ownship};
        \node[aircraft top,fill=red,minimum width=0.4cm] () at (3,1.2) {};
        \node[] () at (4.7,1.2) {\begin{tabular}{l} Intruders included \\ in the state \end{tabular}};
        \node[aircraft top,fill=green,minimum width=0.4cm] () at (3,0.2) {};
        \node[] () at (4.7,0.2) {\begin{tabular}{l} Intruders excluded \\ from the state \end{tabular}};
        \draw[black,thick,dashed] (2.7, -0.8) -- (3.3, -0.8);
        \node[] () at (4.5, -0.8) {\begin{tabular}{l} Sensing range \\ of the ownship \end{tabular}};
        \draw[blue!50,thick,dashed] (2.7, -1.8) -- (3.3, -1.8);
        \node[] () at (4.6, -1.8) {\begin{tabular}{l} Sector division \\ boundaries \end{tabular}};
    \end{tikzpicture}
    \caption{The sensing area equally divided into four sectors. The state is represented by the closest intruder from each sector.}
    \label{fig:sections}
\end{figure}
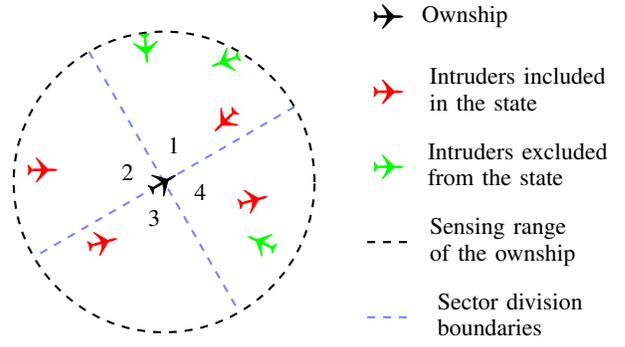

We coarsely model the sensing area of the aircraft as a circle divided into $N$ sectors~\cite{seok2017task}.
The aircraft observes the closest intruder in each sector, forming $N$ pairwise encounters. We then extract $N$ pairwise encounter states, which are referred to as sub-states. The state for the deep correction network is formed by concatenating the $N$ sub-states in the sector ordering. Fig.~\ref{fig:sections} illustrates the sensing area being equally divided into four circular sectors, and the closest intruders in each sector are selected for the state. If an sector has no intruder, then the corresponding sub-state is set to empty (zeros). 

This formulation encodes the approximate spatial locations of the most significant intruders into the state through predefined sector ordering to help the deep neural network better understand the state.

\input{atms2019_li/pics/slices.tex}
\begin{figure}
\begin{tikzpicture}[]
\small
\begin{axis}[height = {2.5in}, legend style = {{at={(0.01, 0.99)}, anchor=north west}}, ylabel = {Alert Frequency}, xlabel = {Number of Intruders}, width = {3.0in}, ymin=0,
xticklabels={},
extra x ticks={1,2,3,4,5,6,7,8},
extra x tick labels={1,2,3,4,5,6,7,8}
]
\addplot+ [mark = {*}, color=magenta,very thick,mark options={fill=magenta}]coordinates {
(1.0, 0.0881)
(2.0, 0.1668)
(3.0, 0.2289)
(4.0, 0.2716)
(5.0, 0.3247)
(6.0, 0.35)
(7.0, 0.3747)
(8.0, 0.4031)
};
\addlegendentry{CorrectedSector}
\addplot+ [mark = {*}, color=pink,very thick,mark options={fill=pink}]coordinates {
(1.0, 0.0847)
(2.0, 0.1683)
(3.0, 0.2408)
(4.0, 0.291)
(5.0, 0.3409)
(6.0, 0.3682)
(7.0, 0.3974)
(8.0, 0.4257)
};
\addlegendentry{CorrectedClosest}
\addplot+ [mark = {*}, color=blue,very thick,mark options={fill=blue}]coordinates {
(1.0, 0.1588)
(2.0, 0.2926)
(3.0, 0.3948)
(4.0, 0.4768)
(5.0, 0.5522)
(6.0, 0.6126)
(7.0, 0.6479)
(8.0, 0.6853)
};
\addlegendentry{VICASMulti}
\addplot+ [mark = {*}, color=red,very thick,mark options={fill=red}]coordinates {
(1.0, 0.1652)
(2.0, 0.2152)
(3.0, 0.2595)
(4.0, 0.2882)
(5.0, 0.3172)
(6.0, 0.3404)
(7.0, 0.3584)
(8.0, 0.3788)
};
\addlegendentry{VICASClosest}
\end{axis}

\end{tikzpicture}
\caption{The probability of CAS issuing Non-COC advisories for different numbers of intruders.}
\label{fig:sensitivity}
\end{figure}
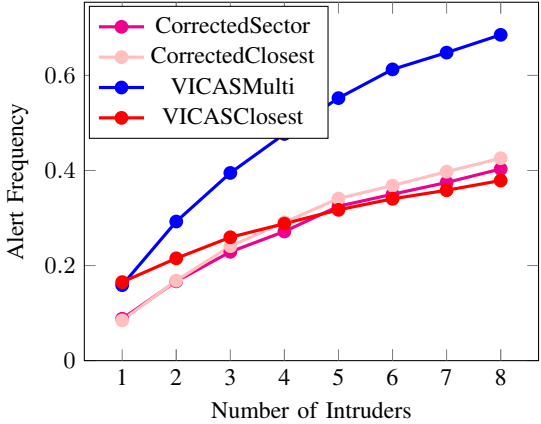
\input{atms2019_li/pics/trajSamples_actionless_.tex}

\subsubsection{Closest Intruders}
Another state formulation method considers the $N$ closest intruders. 
The position, speed, and heading information of the $N$ closest intruders is concatenated into a single observation sorted by their proximity in ascending order.
If there are fewer than $N$ intruders, we take all the existing intruders in to the state and leave the remaining entries of the state empty. 

Though a close distance does not necessarily indicate danger, they are highly related.
This formulation encodes the ordering of threat level into the state, which could also help the deep neural network better understand the state. 
This method has a lower chance of having empty state entries than choosing the closest intruders in $N$ sectors, which could be an advantage during training.

\subsubsection{Destination Information}
We add information about the final destination of the aircraft into the observation to encourage more efficient maneuvering. 
We refer to this additional information as augmented states. 
The augmented states include:
\begin{itemize}
    \item $\theta_\text{dest} \in [-\pi,\pi]$: The angle of the destination relative to the heading of the ownship.
    \item $\rho_\text{dest}$: The distance from the ownship to the destination.
    \item $\rho_\text{dest, prev}$: The distance from the ownship to the destination at previous time step.
\end{itemize}

By including augmented states in the observation, we provide the policy with information that can improve its efficiency.

\subsection{State Transition through Simulation} \label{sec:transition_simulation}
To collect a large amount of experience samples efficiently, a simulator is developed for training the deep correction network. 
The simulator has one learning agent as the ownship. 
Intruders enter the sensing range of the learning agent at angles following the distribution obtained from one million random encounters generated by the Lincoln Laboratory Uncorrelated Encounter Model~\cite{kochenderfer2009airspace}. 

During training, the learning agent follows an $\epsilon$-greedy policy with correction; while the intruders follow a multi-threat resolution policy using the max-min utility decomposition.
We make the intruders' policy stochastic by turning the Q-values into a probability distribution using a softmax function.

The simulation has a predefined episode horizon. In each episode a random destination is given to the ownship. The end of an episode is marked by either the time step reaching the horizon or by the ownship reaching the destination.

\subsection{Reward Function}\label{sec:reward}
The reward function for training the correction network is similar to that of the pairwise conflict resolution MDP. 
The major differences are the reward function for the correction network considers more intruders and it encourages travelling towards the destination. 
For a correction network that considers at most $N$ intruders, its state $s$ can be written as $s = [s_1,\dots,s_N, \theta_\text{dest}, \rho_\text{dest}, \rho_\text{dest, prev}]$.
The reward function is
\begin{equation}
    \begin{aligned}
        R_\text{c}(s, a)
        & = \frac{1}{N} \sum_{i=1}^N \bigg[ -w_\rho \exp\left( \frac{-(\rho(s_i) - \rho_\text{NMAC})}{\rho_\text{NMAC}} \right) \\
        & - w_\text{NMAC} \mathbbm{1}\{\rho(s_i) \leqslant \rho_\text{NMAC}\} \bigg] - w_a \text{turnrate}(a)^2 \\
        & - w_\text{conflict} \mathbbm{1}\{a \neq \text{COC}\} - w_\text{digression}(\rho_\text{dest} - \rho_\text{dest, prev})\\ 
        & - w_\text{deviation}|\theta_\text{dest}| + w_\text{dest} \mathbbm{1}\{\rho_\text{dest} \leqslant \text{DC} \}\text{,}
    \end{aligned}
\end{equation}
where $w_\rho$, $w_\text{NMAC}$, $w_\text{a}$ and $w_\text{conflict}$ have the same purposes as described in \cref{sec:reward} with adjustable values. The parameter $w_\text{digression}$ penalizes the digression of the ownship from the destination in terms of the distance from it to the destination, $w_\text{deviation}$ penalizes the deviation of the ownship's heading from the destination, $w_\text{dest}$ rewards the ownship in reaching its destination (DC is the criterion judging whether the ownship being close enough to the destination). The reward function is constructed this way so that safety and efficiency can be balanced.

\subsection{Corrected Policy}

With the correction network, the corrected policy $\pi_\text{c}$ is extracted
\begin{equation}
    \pi_\text{c}(s) = \argmax_a \left[(1-w_\text{c}) Q_\text{lo}(s,a) + w_\text{c} \delta(s,a)\right]\text{,}
\end{equation}
where $Q_\text{lo}$ is obtained from utility decomposition.

\section{Results}\label{sec:results}
This section compares the safety and efficiency of the following systems: 
\begin{itemize}
    \item \textbf{VICASMulti}: A baseline method based on the CAS computed using value iteration (VICAS) that focuses on pairwise conflict resolution, resolving multi-threat conflicts with max-min utility decomposition.
    \item \textbf{VICASClosest}: A baseline method similar to VICASMulti, resolving multi-threat conflicts by considering the closest intruder.
    \item \textbf{CorrectedSector}: Using VICASMulti as the low-fidelity policy, adding the correction term with state space based on the closest intruders in four circular sectors.
    \item \textbf{CorrectedClosest}: Using VICASMulti as the low-fidelity policy, adding the correction term with state space based on the four closest intruders. 
\end{itemize}
The sensing range of the aircraft is \SI{1000}{\meter} and the NMAC range is defined to be \SI{150}{\meter}.

\subsection{Policy Slices Visualization and Policy Sensitivity}

One intuitive way of understanding what effects the correction term has on the low-fidelity policy is through visualization.
\cref{fig:slices} shows policy slices of different CAS in a four-aircraft encounter.
Headings of all the aircraft as well as the positions of the ownship and the two fixed intruders are fixed. 
The position of the free intruder can be anywhere on the heat map. 
The heat maps show the advisory the CAS would issue to the ownship in response to the position of the free intruder.  

Comparing the policy slices, the alert (non-COC) area of the ownship varies among different CAS. 
The general effects of the correction term are shaping the alert area more compact and more likely to issue COC. 
If a CAS is too sensitive, it could issue advisories too frequent. For example in the first row of the policy slices in \cref{fig:slices}, in an encounter situation where CorrectedClosest issues COC, VICASMulti issues WR instead. Intuitively, early responses are desirable. However, in a dense airspace, a more winding path means higher chances of encountering more intruders. Being less sensitive does not necessarily imply being less safe. When the intruders get closer, for example shown in the second row of the policy slices in \cref{fig:slices}, CorrectedClosest can still issue strong advisories to avoid the threats. A qualitative conclusion we may draw from this is the corrected CAS are less sensitive than the baselines, but still sensitive enough to remain safe. The corrected CAS show higher efficiency in terms of the alert frequency.

To quantify the sensitivity of the CAS, the alert frequency of each CAS given various number of intruders is estimated through a sampling based approach. \cref{fig:sensitivity} shows that the corrected CAS have lower sensitivity than the risk-averse VICASMulti, and have similar sensitivity with the VICASClosest, which only considers the closest intruder.

\subsection{Trajectory Samples}

Sample trajectories of a three-aircraft encounter are visualized in \cref{fig:traj}. 
The encounter is constructed so that when there is no CAS, an NMAC is inevitable. 
VICASMulti has the the maximum minimum distance between the aircraft with the most winding paths. 
The corrected CAS, on the other hand, produce the most efficient trajectories with fewer maneuvers.  
Qualitatively, the example illustrates that the correction term improves the efficiency of the the CAS, while maintaining safety.  

\subsection{Safety and Efficiency Evaluation through Airspace Simulations}
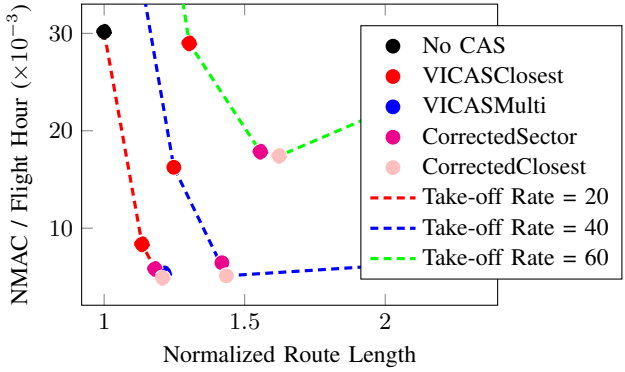
\begin{figure}
\begin{tikzpicture}[]
\small
\begin{axis}[height = {2.2in}, legend style = {{at={(1.3, 0.93)}, anchor=north east}}, ylabel = {NMAC / Flight Hour ($\times 10^{-3}$)}, xmin = {0.92}, xmax = {2.4}, ymax = {33}, xlabel = {Normalized Route Length}, y tick label style={/pgf/number format/fixed}, x tick label style={/pgf/number format/fixed}, width = {2.8in}]\addplot+[draw=none, mark size = {3}, white,mark=*,mark options={fill=black}] coordinates {
(1.0, 57.57)
};
\addlegendentry{No CAS}
\addplot+[draw=none, mark size = {3}, white,mark=*,mark options={fill=red}] coordinates {
(1.248, 16.25)
};
\addlegendentry{VICASClosest}
\addplot+[draw=none, mark size = {3}, white,mark=*,mark options={fill=blue}] coordinates {
(3.088, 8.18)
};
\addlegendentry{VICASMulti}
\addplot+[draw=none, mark size = {3}, white,mark=*,mark options={fill=magenta}] coordinates {
(1.419, 6.45)
};
\addlegendentry{CorrectedSector}
\addplot+[draw=none, mark size = {3}, white,mark=*,mark options={fill=pink}] coordinates {
(1.435, 5.11)
};
\addlegendentry{CorrectedClosest}
\addplot+ [mark = {none}, very thick, color=red]coordinates {
(1.0, 30.16)
(1.135, 8.35)
(1.181, 5.82)
(1.208, 4.9)
};
\addlegendentry{Take-off Rate = 20}
\addplot+[draw=none, mark size = {3}, forget plot,white,mark=*,mark options={fill=black}] coordinates {
(1.0, 30.16)
};
\addplot+[draw=none, mark size = {3}, forget plot,white,mark=*,mark options={fill=red}] coordinates {
(1.135, 8.35)
};
\addplot+[draw=none, mark size = {3}, forget plot,white,mark=*,mark options={fill=blue}] coordinates {
(1.214, 5.35)
};
\addplot+[draw=none, mark size = {3}, forget plot,white,mark=*,mark options={fill=magenta}] coordinates {
(1.181, 5.82)
};
\addplot+[draw=none, mark size = {3}, forget plot,white,mark=*,mark options={fill=pink}] coordinates {
(1.208, 4.9)
};
\addplot+ [mark = {none}, very thick, color=blue]coordinates {
(1.0, 57.57)
(1.248, 16.25)
(1.435, 5.11)
(3.088, 8.18)
};
\addlegendentry{Take-off Rate = 40}
\addplot+[draw=none, mark size = {3}, forget plot,white,mark=*,mark options={fill=black}] coordinates {
(1.0, 85.52)
};
\addplot+[draw=none, mark size = {3}, forget plot,white,mark=*,mark options={fill=red}] coordinates {
(1.303, 28.97)
};
\addplot+[draw=none, mark size = {3}, forget plot,white,mark=*,mark options={fill=blue}] coordinates {
(7.686, 92.32)
};
\addplot+[draw=none, mark size = {3}, forget plot,white,mark=*,mark options={fill=magenta}] coordinates {
(1.556, 17.85)
};
\addplot+[draw=none, mark size = {3}, forget plot,white,mark=*,mark options={fill=pink}] coordinates {
(1.623, 17.42)
};
\addplot+ [mark = {none}, very thick, color=green]coordinates {
(1.0, 85.52)
(1.303, 28.97)
(1.556, 17.85)
(1.623, 17.42)
(7.686, 92.32)
};
\addlegendentry{Take-off Rate = 60}
\end{axis}

\end{tikzpicture}

\caption{Pareto frontiers between safety and efficiency. The unit for take-off rates is flight / km$^2$-hr.}
\label{fig:pareto_takeoffrates}

\end{figure}
We evaluate the safety and efficiency of the CAS in the airspace where aircraft take off at various rates. It is a more challenging scenario than the fixed-number encounters. 

Airspace simulations are run in a 10 km $\times$ 10 km airspace. The initial positions and destinations of the aircraft are uniformly sampled in the airspace. Each simulation runs for \SI{5000}{\second}.

\begin{table*}[t]
    \centering
    \small
    \begin{threeparttable}
    \begin{tabular}{c|l|l|l|l|l|l}
        \hline
        \multirow{2}{*}{Metrics} & \multirow{2}{*}{CAS} & \multicolumn{5}{c}{Take-off Rates (flight / km$^2$-hr)} \\ \cline{3-7}
                                & &5  &10 &20  &40  &60 \\ \hline
        \multirow{5}{*}{\shortstack{NMACs / Flight Hour \\ ($\times 10^{-3}$)}}
                                            & No CAS            &12.07$\pm$0.82 &16.57$\pm$0.40 &30.16$\pm$0.39 &57.57$\pm$0.82 &85.52$\pm$0.85 \\ \cdashline{2-7}
                                            & VICASMulti        &4.83$\pm$0.08 &4.92$\pm$0.25 &5.35$\pm$0.17  &8.18$\pm$2.74  &92.32$\pm$6.36 \\
                                            & VICASClosest      &4.97$\pm$0.21 &6.17$\pm$0.15 &8.35$\pm$0.22 &16.25$\pm$0.30 &28.97$\pm$0.57 \\
                                            & CorrectedSector   &4.45$\pm$0.24 &5.77$\pm$0.26 &5.82$\pm$0.11 &6.45$\pm$0.10 &17.85$\pm$0.72 \\
                                            & CorrectedClosest   &\textbf{2.99$\pm$0.07} &\textbf{4.72$\pm$0.14}  &\textbf{4.90$\pm$0.34} &\textbf{5.11$\pm$0.12}  &\textbf{17.42$\pm$0.35} \\ \hline
        
        \multirow{5}{*}{\shortstack{Normalized Route Length}}
                                            & No CAS            &1.0 &1.0 &1.0 &1.0 &1.0 \\ \cdashline{2-7}
                                            & VICASMulti     &1.109$\pm$0.006 &1.113$\pm$0.007 &1.214$\pm$0.006 &3.088$\pm$0.191 &7.686$\pm$0.601 \\
                                            & VICASClosest   &1.092$\pm$0.003 &1.101$\pm$0.009 &1.135$\pm$0.001 &1.248$\pm$0.004 &1.303$\pm$0.003 \\
                                            & CorrectedSector   &1.095$\pm$0.002 &1.114$\pm$0.008 &1.181$\pm$0.003 &1.419$\pm$0.003 &1.556$\pm$0.008  \\
                                            & CorrectedClosest  &1.108$\pm$0.003 &1.140$\pm$0.008 &1.208$\pm$0.006 &1.435$\pm$0.003 &1.623$\pm$0.005  \\ \hline        
        \multirow{5}{*}{$D_\text{TV}$ ($\times 10^{-2}$)}
                                            & No CAS (ref)        &- &- &- &- &- \\ \cdashline{2-7}
                                            & VICASMulti     &3.142$\pm$0.107 &3.407$\pm$0.091 &6.139$\pm$0.063 &30.638$\pm$0.034 &41.389$\pm$0.189 \\
                                            & VICASClosest   &1.106$\pm$0.117 &1.660$\pm$0.087 &2.756$\pm$0.066 &5.749$\pm$0.049 &7.411$\pm$0.086  \\
                                            & CorrectedSector  &1.993$\pm$0.129 &3.070$\pm$0.113 &4.943$\pm$0.064 &16.303$\pm$0.080 &16.383$\pm$0.046  \\
                                            & CorrectedClosest  &2.416$\pm$0.109 &2.923$\pm$0.085 &5.933$\pm$0.067 &21.064$\pm$0.078 &28.121$\pm$0.074  \\ \hline     
    \end{tabular}
    \end{threeparttable}  
    \caption{Performance metrics (as mean $\pm$ standard error) for different CAS and taking-off rates.}
    \label{tab:performance}
\end{table*}

\subsubsection{Safety}\label{sec:results:safety}
Safety is measured by number of NMACs per flight hour. \cref{tab:performance} shows that the NMAC rates of CorrectedClosest are the lowest for all take-off rates. Note that VICASMuilti performs well until the take-off rate exceeds 40 flight / km$^2$-hr, where the NMAC rates increase dramatically.

\subsubsection{Efficiency}\label{sec:results:efficiency}
Efficiency is measured by the ratio between the length of the actual taken path and the nominal distance from the start to the destination, i.e. the normalized route length. Listed in \cref{tab:performance}, the normalized route length of VICASMulti dramatically increases when the take-off rate exceeds 40 flight / km$^2$-hr. This explains the increase in the NMAC rates. As aircraft follow overly winding paths generated by VICASMulti, the chance of NMAC increases. VICASClosest has the lowest the normalized route length at an expense of safety. The normalized route lengths of the corrected CAS increase at a rates proportional to the increase in the taking off rates, while maintaining low NMAC rates. This indicates that unlike VICASClosest, the corrected CAS are able to issue necessary advisories to stay safe; and unlike VICASMulti, efficiency is also considered at high take-off rates. We may say the safety and efficiency of the corrected CAS are balanced.

\subsubsection{Safety versus Efficiency Trade-off}
\cref{fig:pareto_takeoffrates} shows the trade-off between safety and efficiency. Pareto frontiers are plotted for different take-off rates. We can see that the corrected CAS have the lowest NMAC rates with low normalized route lengths. The corrected CAS are corresponding with the best performing points lie at the bottom left corners of the Pareto frontiers.

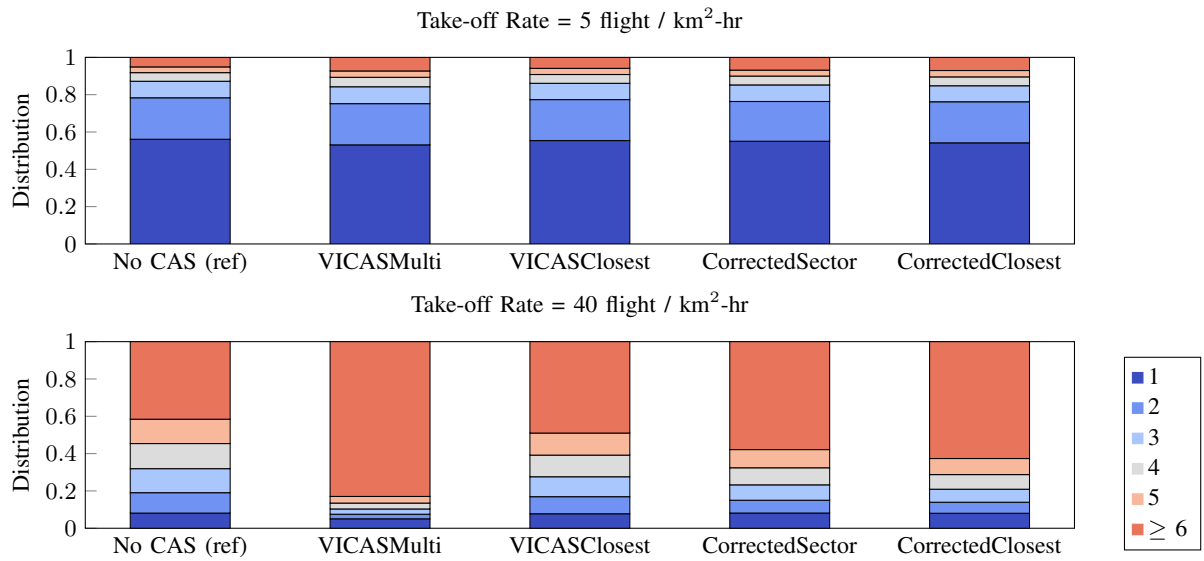
\begin{figure*}[t]
\centering
\begin{tikzpicture}[]
\small
\definecolor{color0}{rgb}{0.2298057,0.298717966,0.753683153}
\definecolor{color1}{rgb}{0.435814806305882,0.570707303152941,0.951717381282353}
\definecolor{color2}{rgb}{0.667252924333333,0.779176457,0.992959213}
\definecolor{color3}{rgb}{0.867427635086275,0.864376599772549,0.862602462019608}
\definecolor{color4}{rgb}{0.968203399,0.7208441,0.612292991333333}
\definecolor{color5}{rgb}{0.905783478011765,0.455185692164706,0.355335883847059}

\begin{groupplot}[group style={vertical sep = 1.3cm, group size=1 by 2}]
\nextgroupplot
[
legend cell align={left},
legend style={at={(1.05,0.4)}, anchor=west},
tick align=inside,
tick pos=left,
x grid style={white!69.01960784313725!black},
xmin=-0.475, xmax=4.475,
xtick={0,1,2,3,4},
xticklabels={No CAS (ref),VICASMulti,VICASClosest,CorrectedSector,CorrectedClosest},
y grid style={white!69.01960784313725!black},
ymin=0, ymax=1, 
width={5.8in}, height={1.6in},
ylabel={Distribution},
title={Take-off Rate = 5 flight / km$^2$-hr}
]
\draw[fill=color0,draw=black] (axis cs:-0.25,0) rectangle (axis cs:0.25,0.561053);
\draw[fill=color0,draw=black] (axis cs:0.75,0) rectangle (axis cs:1.25,0.531042);
\draw[fill=color0,draw=black] (axis cs:1.75,0) rectangle (axis cs:2.25,0.55409);
\draw[fill=color0,draw=black] (axis cs:2.75,0) rectangle (axis cs:3.25,0.550466);
\draw[fill=color0,draw=black] (axis cs:3.75,0) rectangle (axis cs:4.25,0.541919);
\draw[fill=color1,draw=black] (axis cs:-0.25,0.561053) rectangle (axis cs:0.25,0.783429);
\draw[fill=color1,draw=black] (axis cs:0.75,0.531042) rectangle (axis cs:1.25,0.752013);
\draw[fill=color1,draw=black] (axis cs:1.75,0.55409) rectangle (axis cs:2.25,0.773693);
\draw[fill=color1,draw=black] (axis cs:2.75,0.550466) rectangle (axis cs:3.25,0.7635);
\draw[fill=color1,draw=black] (axis cs:3.75,0.541919) rectangle (axis cs:4.25,0.762009);
\draw[fill=color2,draw=black] (axis cs:-0.25,0.783429) rectangle (axis cs:0.25,0.8720347);
\draw[fill=color2,draw=black] (axis cs:0.75,0.752013) rectangle (axis cs:1.25,0.8424721);
\draw[fill=color2,draw=black] (axis cs:1.75,0.773693) rectangle (axis cs:2.25,0.8609763);
\draw[fill=color2,draw=black] (axis cs:2.75,0.7635) rectangle (axis cs:3.25,0.8523232);
\draw[fill=color2,draw=black] (axis cs:3.75,0.762009) rectangle (axis cs:4.25,0.8478758);
\draw[fill=color3,draw=black] (axis cs:-0.25,0.8720347) rectangle (axis cs:0.25,0.9181094);
\draw[fill=color3,draw=black] (axis cs:0.75,0.8424721) rectangle (axis cs:1.25,0.8931682);
\draw[fill=color3,draw=black] (axis cs:1.75,0.8609763) rectangle (axis cs:2.25,0.908122);
\draw[fill=color3,draw=black] (axis cs:2.75,0.8523232) rectangle (axis cs:3.25,0.8998863);
\draw[fill=color3,draw=black] (axis cs:3.75,0.8478758) rectangle (axis cs:4.25,0.8953579);
\draw[fill=color4,draw=black] (axis cs:-0.25,0.9181094) rectangle (axis cs:0.25,0.948829);
\draw[fill=color4,draw=black] (axis cs:0.75,0.8931682) rectangle (axis cs:1.25,0.927122);
\draw[fill=color4,draw=black] (axis cs:1.75,0.908122) rectangle (axis cs:2.25,0.941155);
\draw[fill=color4,draw=black] (axis cs:2.75,0.8998863) rectangle (axis cs:3.25,0.9316222);
\draw[fill=color4,draw=black] (axis cs:3.75,0.8953579) rectangle (axis cs:4.25,0.9295197);
\draw[fill=color5,draw=black] (axis cs:-0.25,0.948829) rectangle (axis cs:0.25,1.0000002);
\draw[fill=color5,draw=black] (axis cs:0.75,0.927122) rectangle (axis cs:1.25,0.9999998);
\draw[fill=color5,draw=black] (axis cs:1.75,0.941155) rectangle (axis cs:2.25,0.9999999);
\draw[fill=color5,draw=black] (axis cs:2.75,0.9316222) rectangle (axis cs:3.25,1.0000001);
\draw[fill=color5,draw=black] (axis cs:3.75,0.9295197) rectangle (axis cs:4.25,0.9999999);
\legend{};

\nextgroupplot
[
legend cell align={left},
legend entries={{1},{2},{3},{4},{5},{$\ge$ 6}},
legend style={at={(1.05,0.4)}, anchor=west},
tick align=inside,
tick pos=left,
x grid style={white!69.01960784313725!black},
xmin=-0.475, xmax=4.475,
xtick={0,1,2,3,4},
xticklabels={No CAS (ref),VICASMulti,VICASClosest,CorrectedSector,CorrectedClosest},
y grid style={white!69.01960784313725!black},
ymin=0, ymax=1, 
width={5.8in}, height={1.6in},
ylabel={Distribution},
title={Take-off Rate = 40 flight / km$^2$-hr}
]

\addlegendimage{only marks,mark=square*,color=color0}
\addlegendimage{only marks,mark=square*,color=color1}
\addlegendimage{only marks,mark=square*,color=color2}
\addlegendimage{only marks,mark=square*,color=color3}
\addlegendimage{only marks,mark=square*,color=color4}
\addlegendimage{only marks,mark=square*,color=color5}

\draw[fill=color0,draw=black] (axis cs:-0.25,0) rectangle (axis cs:0.25,0.0812906);
\draw[fill=color0,draw=black] (axis cs:0.75,0) rectangle (axis cs:1.25,0.0506228);
\draw[fill=color0,draw=black] (axis cs:1.75,0) rectangle (axis cs:2.25,0.0776443);
\draw[fill=color0,draw=black] (axis cs:2.75,0) rectangle (axis cs:3.25,0.0815799);
\draw[fill=color0,draw=black] (axis cs:3.75,0) rectangle (axis cs:4.25,0.0803869);
\draw[fill=color1,draw=black] (axis cs:-0.25,0.0812906) rectangle (axis cs:0.25,0.1906046);
\draw[fill=color1,draw=black] (axis cs:0.75,0.0506228) rectangle (axis cs:1.25,0.0751425);
\draw[fill=color1,draw=black] (axis cs:1.75,0.0776443) rectangle (axis cs:2.25,0.1691425);
\draw[fill=color1,draw=black] (axis cs:2.75,0.0815799) rectangle (axis cs:3.25,0.1501912);
\draw[fill=color1,draw=black] (axis cs:3.75,0.0803869) rectangle (axis cs:4.25,0.1394436);
\draw[fill=color2,draw=black] (axis cs:-0.25,0.1906046) rectangle (axis cs:0.25,0.3193606);
\draw[fill=color2,draw=black] (axis cs:0.75,0.0751425) rectangle (axis cs:1.25,0.1034045);
\draw[fill=color2,draw=black] (axis cs:1.75,0.1691425) rectangle (axis cs:2.25,0.2759615);
\draw[fill=color2,draw=black] (axis cs:2.75,0.1501912) rectangle (axis cs:3.25,0.2325994);
\draw[fill=color2,draw=black] (axis cs:3.75,0.1394436) rectangle (axis cs:4.25,0.2092375);
\draw[fill=color3,draw=black] (axis cs:-0.25,0.3193606) rectangle (axis cs:0.25,0.4538886);
\draw[fill=color3,draw=black] (axis cs:0.75,0.1034045) rectangle (axis cs:1.25,0.1348732);
\draw[fill=color3,draw=black] (axis cs:1.75,0.2759615) rectangle (axis cs:2.25,0.3919585);
\draw[fill=color3,draw=black] (axis cs:2.75,0.2325994) rectangle (axis cs:3.25,0.324061);
\draw[fill=color3,draw=black] (axis cs:3.75,0.2092375) rectangle (axis cs:4.25,0.2875897);
\draw[fill=color4,draw=black] (axis cs:-0.25,0.4538886) rectangle (axis cs:0.25,0.5843626);
\draw[fill=color4,draw=black] (axis cs:0.75,0.1348732) rectangle (axis cs:1.25,0.1704685);
\draw[fill=color4,draw=black] (axis cs:1.75,0.3919585) rectangle (axis cs:2.25,0.5102505);
\draw[fill=color4,draw=black] (axis cs:2.75,0.324061) rectangle (axis cs:3.25,0.4216188);
\draw[fill=color4,draw=black] (axis cs:3.75,0.2875897) rectangle (axis cs:4.25,0.3737218);
\draw[fill=color5,draw=black] (axis cs:-0.25,0.5843626) rectangle (axis cs:0.25,0.9999996);
\draw[fill=color5,draw=black] (axis cs:0.75,0.1704685) rectangle (axis cs:1.25,1.0000005);
\draw[fill=color5,draw=black] (axis cs:1.75,0.5102505) rectangle (axis cs:2.25,0.9999995);
\draw[fill=color5,draw=black] (axis cs:2.75,0.4216188) rectangle (axis cs:3.25,0.9999998);
\draw[fill=color5,draw=black] (axis cs:3.75,0.3737218) rectangle (axis cs:4.25,0.9999998);

\end{groupplot}
\end{tikzpicture}
\caption{The encounter distribution for different take-off rates.}
\label{fig:distribution}
\end{figure*}
\begin{figure}
\begin{tikzpicture}[]
\small
\begin{axis}[height = {2.4in}, legend style = {{at={(0.01, 0.99)}, anchor=north west}}, ylabel = {Probability of NMAC}, xlabel = {Number of Aircraft}, width = {3.0in}, ymax=0.032,ymin=-0.001,
xticklabels={},
extra x ticks={2,3,4,5,6,7},
extra x tick labels={2,3,4,5,6,7}]
\addplot+ [mark=none, color=magenta,very thick,mark options={fill=magenta}, error bars/.cd, 
x dir=both, x explicit, y dir=both, y explicit]
table [
x error plus=ex+, x error minus=ex-, y error plus=ey+, y error minus=ey-
] {
x y ex+ ex- ey+ ey-
2.0 0.00050 0 0 0.00050 0.00050
3.0 0.00200 0 0 0.00100 0.00100
4.0 0.00380 0 0 0.00158 0.001528
5.0 0.00620 0 0 0.00228 0.00228
6.0 0.01600 0 0 0.00307 0.00307
7.0 0.02600 0 0 0.00325 0.00325
};
\addlegendentry{correctedSector}
\addplot+ [mark = none, color=pink,very thick,mark options={fill=pink}, error bars/.cd, 
x dir=both, x explicit, y dir=both, y explicit]
table [
x error plus=ex+, x error minus=ex-, y error plus=ey+, y error minus=ey-
] {
x y ex+ ex- ey+ ey-
2.0 0.0 0 0 0 0
3.0 0.0 0 0 0 0
4.0 0.001 0 0 0.00070 0.00070
5.0 0.002 0 0 0.00100 0.00100
6.0 0.006 0 0 0.00173 0.00173
7.0 0.0115 0 0 0.00246 0.00246
};
\addlegendentry{correctedClosest}
\addplot+ [mark=none, color=blue,very thick,mark options={fill=blue}, error bars/.cd, 
x dir=both, x explicit, y dir=both, y explicit]
table [
x error plus=ex+, x error minus=ex-, y error plus=ey+, y error minus=ey-
] {
x y ex+ ex- ey+ ey-
2.0 0.00200 0 0 0.00100 0.00100
3.0 0.00800 0 0 0.00199 0.00199
4.0 0.00950 0 0 0.00205 0.00205
5.0 0.01450 0 0 0.00267 0.00267
6.0 0.02150 0 0 0.00324 0.00324
7.0 0.02500 0 0 0.00349 0.00349
};
\addlegendentry{VICASMulti}
\addplot+ [mark=none, color=red,very thick,mark options={fill=red}, error bars/.cd, 
x dir=both, x explicit, y dir=both, y explicit]
table [
x error plus=ex+, x error minus=ex-, y error plus=ey+, y error minus=ey-
] {
x y ex+ ex- ey+ ey-
2.0 0.00300 0 0 0.00122 0.00122
3.0 0.02700 0 0 0.00363 0.00363
4.0 0.13600 0 0 0.00767 0.00767
5.0 0.28775 0 0 0.01018 0.01018
6.0 0.48200 0 0 0.01127 0.01127
7.0 0.69200 0 0 0.01096 0.01096
};
\addlegendentry{VICASClosest}
\legend{};
\end{axis}

\end{tikzpicture}
\caption{Probability of NMAC when resolving conflicts with different number of aircraft.}
\label{fig:resolve_nmac}
\end{figure}
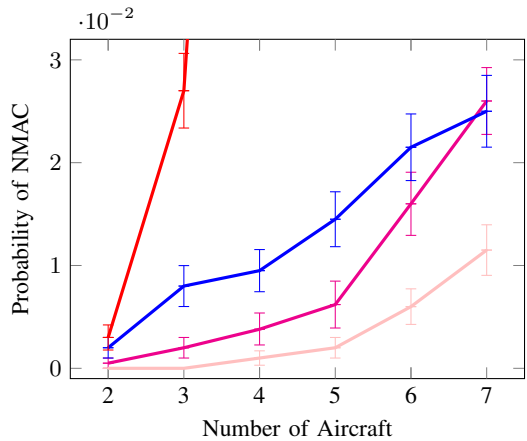

\subsubsection{Impact on Encounter Distribution}
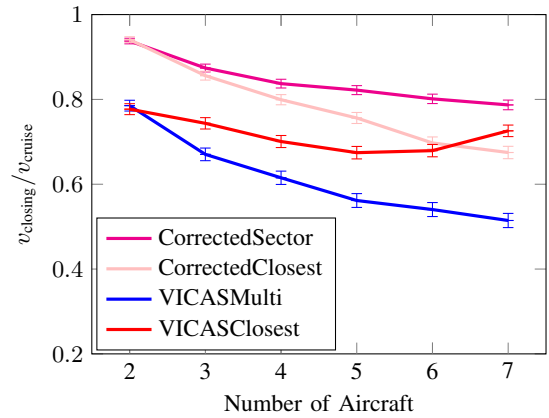
\begin{figure}
\begin{tikzpicture}[]
\small
\begin{axis}[height = {2.4in}, legend style = {{at={(0.01, 0.01)}, anchor=south west}}, ylabel = {$v_\text{closing}/v_\text{cruise}$}, xlabel = {Number of Aircraft}, width = {3.0in}, ymax=1.0, ymin=0.2,
xticklabels={},
extra x ticks={2,3,4,5,6,7},
extra x tick labels={2,3,4,5,6,7}]
\addplot+ [mark = none, color=magenta, very thick,mark options={fill=magenta}, error bars/.cd, 
x dir=both, x explicit, y dir=both, y explicit]
table [
x error plus=ex+, x error minus=ex-, y error plus=ey+, y error minus=ey-
] {
x y ex+ ex- ey+ ey-
2.0 0.9376 0 0 0.0066 0.0066
3.0 0.8740 0 0 0.0092 0.0092
4.0 0.8372 0 0 0.0103 0.0103
5.0 0.8219 0 0 0.0106 0.0106
6.0 0.8013 0 0 0.0111 0.0111
7.0 0.7870 0 0 0.0114 0.0114
};
\addlegendentry{CorrectedSector}
\addplot+ [mark = none, color=pink,very thick,mark options={fill=pink}, error bars/.cd, 
x dir=both, x explicit, y dir=both, y explicit]
table [
x error plus=ex+, x error minus=ex-, y error plus=ey+, y error minus=ey-
] {
x y ex+ ex- ey+ ey-
2.0 0.9409 0 0 0.0069 0.0069
3.0 0.8560 0 0 0.0104 0.0104
4.0 0.7993 0 0 0.0120 0.0120
5.0 0.7562 0 0 0.0130 0.0130
6.0 0.6973 0 0 0.0142 0.0142
7.0 0.6748 0 0 0.0146 0.0146
};
\addlegendentry{CorrectedClosest}
\addplot+ [mark=none, color=blue,very thick,mark options={fill=blue}, error bars/.cd, 
x dir=both, x explicit, y dir=both, y explicit]
table [
x error plus=ex+, x error minus=ex-, y error plus=ey+, y error minus=ey-
] {
x y ex+ ex- ey+ ey-
2.0 0.7853 0 0 0.0127 0.0127
3.0 0.6704 0 0 0.0149 0.0149
4.0 0.6152 0 0 0.0157 0.0157
5.0 0.5616 0 0 0.0163 0.0163
6.0 0.5403 0 0 0.0165 0.0165
7.0 0.5145 0 0 0.0167 0.0167
};
\addlegendentry{VICASMulti}
\addplot+ [mark = none, color=red,very thick,mark options={fill=red}, error bars/.cd, 
x dir=both, x explicit, y dir=both, y explicit]
table [
x error plus=ex+, x error minus=ex-, y error plus=ey+, y error minus=ey-
] {
x y ex+ ex- ey+ ey-
2.0 0.7770 0 0 0.0129 0.0129
3.0 0.7436 0 0 0.0134 0.0134
4.0 0.7007 0 0 0.0142 0.0142
5.0 0.6744 0 0 0.0147 0.0147
6.0 0.6792 0 0 0.0145 0.0145
7.0 0.7259 0 0 0.0136 0.0136
};
\addlegendentry{VICASClosest}
\end{axis}

\end{tikzpicture}
\caption{Closing speed / cruise speed when resolving conflicts with different number of aircraft.}
\label{fig:veff}
\end{figure}

One way to capture the average decision burden on a CAS in a given airspace is to consider the number of intruders in a conflict~\cite{golding2018meterics}.
We extend this notion to a distribution over the average number of intruders in all encounters within the airspace.
This metric, called the encounter distribution, provides a general notion of CAS effectiveness in the airspace, as encounters with more intruders will have a higher likelihood of occurring in a dense airspace and are more likely to result in an NMAC.

We observe that a CAS will be less effective when attempting to resolve a conflict with more than one intruder.
Therefore, we define an airspace to be dense with regards to the encounter distribution.
Formally, a dense airspace is where the expectation of a multi-threat encounter is above some threshold value $\lambda$, $\mathbb{E}[N_\text{intruder} > 1] \geq \lambda$, where $N_\text{intruder}$ is the number of intruders in an encounter. 

\cref{fig:distribution} illustrates the encounter distributions for different take-off rates.
At the take-off rate of 5~flight~/~km$^2$-hr, the encounter distributions are not greatly impacted by the CAS. Over 50\% of the encounters are pairwise. 
At the take-off rate of 40~flight~/~km$^2$-hr, the low efficiency of VICASMulti drives the encounter distribution towards higher numbers of intruders, which increases the average complexity of conflicts in the airspace. 
The corrected CAS impact the encounter distribution more than VICASClosest does, which can again be explained by the fact that the corrected CAS issue advisories more frequently than VICASClosest to stay safe. 

By computing the total variation divergence between the encounter distribution for an airspace without an active CAS and one with an active CAS, we can quantify how a CAS changes the encounter structure of an airspace. 
Namely, given an encounter distribution for an airspace with no active CAS, $P_\text{No}$, and encounter distribution for an airspace with an active CAS, $P_\text{CAS}$, the impact of the active CAS on the airspace is
\begin{equation}
    D_\text{TV}(P_\text{No} \parallel P_\text{CAS}) = \frac{1}{2} \sum_{x \in \Omega} |P_\text{No}(x) - P_\text{CAS}(x)|\text{,}
\end{equation}
where $\Omega$ is the support of the encounter distribution. 
The impact on encounter distribution of a CAS is measured by $D_\text{TV}$. The results are listed in \cref{tab:performance}.






\subsection{Safety and Efficiency Evaluation through Stress Tests}
A stress test is designed to further examine the safety and efficiency of the CAS~\cite{ong2016markov}. In the stress test, the number of aircraft is fixed. They are randomly initialized in an annulus with an inner radius of \SI{2000}{\meter} and an outer radius of \SI{4000}{\meter}. Their initial headings are toward the center of the annulus to make sure the possibility of encountering. 

\cref{fig:resolve_nmac} shows the probability of NMAC given different numbers of aircraft in the stress test. The corrected CAS are safer than the baselines, in which CorrectedClosest performs the best. VICASClosest shows extremely high probability of NMAC when the encounter is more complicated than pairwise.

The ratio between the speed of aircraft getting close to destinations and the cruise speed (speed efficiency, $v_\text{closing}/v_\text{cruise}$) is tracked in the stress tests as an additional indication of efficiency. \cref{fig:veff} shows that the corrected CAS have superior speed efficiency. 

We define the severity of an NMAC as
\begin{equation}
    \text{Severity} = \max\{0, 1 - D_\text{min} / \text{NMAC Range}\}.
\end{equation}
The NMAC severity is tracked in the stress tests. The NMAC severity is not strongly correlated with number of aircraft in an encounter. However, it differs among different CAS. \cref{tab:severe} shows that CorrectedSector has the lowest NMAC severity among all the CAS, whereas VICASMulti has the highest NMAC severity.

\begin{table}[]
    \centering
    \small
    \begin{tabular}{lr}
        \toprule
        CAS  &NMAC Severity \\ \midrule
        No CAS & 0.1894$\pm$0.0043 \\ \hdashline
        VICASMulti & 0.3280$\pm$0.0129 \\
        VICASClosest & 0.2017$\pm$0.0162 \\
        CorrectedSector &  \textbf{0.1701$\pm$0.0090} \\
        CorrectedClosest &  0.2001$\pm$0.0256 \\
        \bottomrule
    \end{tabular}
    \caption{NMAC severity (as mean $\pm$ standard error).}
    \label{tab:severe}
\end{table}

\section{Conclusions and Further Work}
In this paper, we assessed the safety and efficiency of CAS operation in dense airspace.
We found that operating table-based CAS using utility decomposition is effective in low density airspace, but the performance can be further improved in dense airspace.
We applied a correction term trained through deep reinforcement learning on top of the utility decomposition to better approximate an optimal policy for dense airspace.
By adding the correction term, we successfully improved the safety and efficiency of CAS performance in both pairwise and multi-threat encounters.
The correction term led to emergent behavior in which the CAS balanced its awareness of the risk from intruders and the goal of the operation. 
The corrected CAS demonstrated superior safety performance with relatively high efficiency and low impact on the encounter distribution of an airspace.

In the future, we could train CAS for multi-threat conflict resolution using deep reinforcement learning from scratch and try other deep reinforcement learning algorithms, such as trust region policy optimization~\cite{schulman2015trust} and proximal policy optimization~\cite{schulman2017proximal}. 
The relationship between collision avoidance and flight planning could be examined as well, similar to have it has been done in traditional ATM~\cite{radanovic2017self}.
In addition, a more sophisticated aircraft model could be used in future work.

\printbibliography

@String { aaai        = {AAAI Conference on Artificial Intelligence (AAAI)} }

@String { aamas       = {International Conference on Autonomous Agents and Multiagent Systems (AAMAS)} }

@String { acc         = {American Control Conference (ACC)} }

@String { dasc        = {Digital Avionics Systems Conference (DASC)} }

@String { icml        = {International Conference on Machine Learning (ICML)} }

@String { icra        = {IEEE International Conference on Robotics and Automation (ICRA)} }

@String { iros        = {IEEE/RSJ International Conference on Intelligent Robots and Systems (IROS)} }

@String { jgcd        = {AIAA Journal of Guidance, Control, and Dynamics} }

@String { mit         = {Massachusetts Institute of Technology} }

@String { atm         = {USA/Europe Air Traffic Management Research and Development Seminar}}

@article{kochenderfer2012next,
  Title = {Next generation airborne collision avoidance system},
Author = {Mykel J. Kochenderfer and Jessica E. Holland and James P. Chryssanthacopoulos},
Journal = {Lincoln Laboratory Journal},
Year = {2012},
Number = {1},
Pages = {17-33},
Volume = {19}
}

@inproceedings{ong2015short,
  Title = {Short-term conflict resolution for unmanned aircraft traffic management},
Author = {Hao Yi Ong and Mykel J. Kochenderfer},
Booktitle = dasc,
Year = {2015},
Doi = {10.1109/DASC.2015.7311424}
}

@inproceedings{julian2016policy,
  author = {Kyle Julian and Jessica Lopez and Jeffrey S. Brush and Michael Owen and Mykel J. Kochenderfer},
title = {Policy compression for aircraft collision avoidance systems},
booktitle = dasc,
year = {2016},
doi = {10.1109/DASC.2016.7778091},
}

@TechReport{kochenderfer2011robust,
author = {Mykel J. Kochenderfer and James P. Chryssanthacopoulos},
title = {Robust airborne collision avoidance through dynamic programming},
institution = {Massachusetts Institute of Technology, Lincoln Laboratory},
year = {2011},
type = {Project Report},
number = {ATC-371},
}

@techreport{davies2018comparative,
  title={Comparative Analysis of {ACAS} {X}u and {DAIDALUS} Detect-and-Avoid Systems},
  author={Davies, Jason T and Wu, Minghong G},
  year={2018},
  number = {NASA/TM-2018-219773},
  type = {Technical Memorandum},
  institution={National Aeronautics and Space Administration}
}

@article{bouton2018utility,
  title={Utility Decomposition with Deep Corrections for Scalable Planning under Uncertainty},
  author={Bouton, Maxime and Julian, Kyle and Nakhaei, Alireza and Fujimura, Kikuo and Kochenderfer, Mykel J},
  journal={arXiv preprint arXiv:1802.01772},
  year={2018}
}

@article{mnih2015human,
  title={Human-level control through deep reinforcement learning},
  author={Mnih, Volodymyr and Kavukcuoglu, Koray and Silver, David and Rusu, Andrei A and Veness, Joel and Bellemare, Marc G and Graves, Alex and Riedmiller, Martin and Fidjeland, Andreas K and Ostrovski, Georg and others},
  journal={Nature},
  volume={518},
  number={7540},
  pages={529},
  year={2015},
  publisher={Nature Publishing Group}
}

@book{jenkins2017forecast,
  title={Forecast of the Commercial UAS Package Delivery Market},
  author={Jenkins, Darryl and Vasigh, Bijan and Oster, Clint and Larsen, Tulinda},
  year={2017},
  publisher={Embry-Riddle Aeronautical University}
}

@inproceedings{kochenderfer2009airspace,
  author = {Mykel J. Kochenderfer and Leo P. Espindle and Matthew W. M. Edwards and James K. Kuchar and J. D. Griffith},
title = {Airspace encounter models for conventional and unconventional aircraft},
booktitle = atm,
year = {2009},
}

@book{bellman2015applied,
  title={Applied dynamic programming},
  author={Bellman, Richard E and Dreyfus, Stuart E},
  volume={2050},
  year={2015},
  publisher={Princeton University Press}
}

@book{kochenderfer2015decision,
  Title = {Decision Making Under Uncertainty: Theory and Application},
Author = {Mykel J. Kochenderfer},
Publisher = {MIT Press},
Year = {2015}
 }

@inproceedings{bertuccelli2008robust,
  title={Robust {Markov} decision processes using sigma point sampling},
  author={Bertuccelli, LF and How, JP},
  booktitle=acc,
  pages={5003--5008},
  year={2008},
  organization={IEEE}
}

@article{chryssanthacopoulos2012decomposition,
  title={Decomposition methods for optimized collision avoidance with multiple threats},
  author={Chryssanthacopoulos, James P and Kochenderfer, Mykel J},
  journal={Journal of Guidance, Control, and Dynamics},
  volume={35},
  number={2},
  pages={398--405},
  year={2012}
}

@article{williamson1989development,
  title={Development and operation of the traffic alert and collision avoidance system ({TCAS})},
  author={Williamson, Thomas and Spencer, Ned A},
  journal={Proceedings of the IEEE},
  volume={77},
  number={11},
  pages={1735--1744},
  year={1989},
  publisher={IEEE}
}

@article{karthik2018blueprint,
  title={Blueprint for the Sky: The Roadmap for the Safe Integration of Autonomous Aircraft},
  author={Balakrishnan, Karthik and Polastre, Joe and Mooberry, Jessie and Golding, Richard and Sachs, Peter},
  journal={Airbus UTM, San Francisco, CA},
  year={2018}
}

@inproceedings{mueller2016multi,
  author = {Eric Mueller and Mykel J. Kochenderfer},
title = {Multi-rotor aircraft collision avoidance using partially observable {M}arkov decision processes},
booktitle = {AIAA Modeling and Simulation Conference},
year = {2016},
doi = {10.2514/6.2016-3673},
}

@inproceedings{thipphavong2017ensuring,
  title={Ensuring Interoperability between UAS Detect-and-Avoid and Manned Aircraft Collision Avoidance},
  author={Thipphavong, David and Cone, Andrew and Lee, Seung Man and Santiago, Confesor},
  booktitle={USA/Europe Air Traffic Management Research and Development Seminar},
  year={2017}
}

@inproceedings{mcfadyen2013aircraft,
  title={Aircraft collision avoidance using spherical visual predictive control and single point features},
  author={Mcfadyen, Aaron and Mejias, Luis and Corke, Peter and Pradalier, C{\'e}dric},
  booktitle=iros,
  pages={50--56},
  year={2013},
  organization={IEEE}
}

@inproceedings{manfredi2016introduction,
  title={An introduction to {ACAS} {X}u and the challenges ahead},
  author={Manfredi, Guido and Jestin, Yannick},
  booktitle=dasc,
  pages={1--9},
  year={2016},
  organization={IEEE}
}

@incollection{van2011reciprocal,
  title={Reciprocal n-body collision avoidance},
  author={Van Den Berg, Jur and Guy, Stephen J and Lin, Ming and Manocha, Dinesh},
  booktitle={Robotics Research},
  pages={3--19},
  year={2011},
  publisher={Springer}
}

@article{mukhtar2015vehicle,
  title={Vehicle Detection Techniques for Collision Avoidance Systems: a Review.},
  author={Mukhtar, Amir and Xia, Likun and Tang, Tong Boon},
  journal={IEEE Transactions on Intelligent Transportation Systems},
  volume={16},
  number={5},
  pages={2318--2338},
  year={2015}
}

@inproceedings{chen2015decoupled,
  title={Decoupled multiagent path planning via incremental sequential convex programming},
  author={Chen, Yufan and Cutler, Mark and How, Jonathan P},
  booktitle=icra,
  pages={5954--5961},
  year={2015},
  organization={IEEE}
}

@inproceedings{tang2015mixed,
  title={Mixed integer quadratic program trajectory generation for a quadrotor with a cable-suspended payload},
  author={Tang, Sarah and Kumar, Vijay},
  booktitle=icra,
  pages={2216--2222},
  year={2015},
  organization={IEEE}
}

@article{foerster2017stabilising,
  title={Stabilising experience replay for deep multi-agent reinforcement learning},
  author={Foerster, Jakob and Nardelli, Nantas and Farquhar, Gregory and Afouras, Triantafyllos and Torr, Philip HS and Kohli, Pushmeet and Whiteson, Shimon},
  journal={arXiv preprint arXiv:1702.08887},
  year={2017}
}

@inproceedings{foerster2018counterfactual,
  title={Counterfactual multi-agent policy gradients},
  author={Foerster, Jakob N and Farquhar, Gregory and Afouras, Triantafyllos and Nardelli, Nantas and Whiteson, Shimon},
  booktitle=aaai,
  year={2018}
}

@inproceedings{gupta2017cooperative,
  author = {Jayesh K. Gupta and Maxim Egorov and Mykel J. Kochenderfer},
title = {Cooperative multi-agent control using deep reinforcement learning},
booktitle = {Adaptive Learning Agents Workshop, } # aamas,
year = {2017},
doi = {10.1007/978-3-319-71682-4_5},
}

@inproceedings{chen2017decentralized,
  title={Decentralized non-communicating multiagent collision avoidance with deep reinforcement learning},
  author={Chen, Yu Fan and Liu, Miao and Everett, Michael and How, Jonathan P},
  booktitle=icra,
  pages={285--292},
  year={2017},
  organization={IEEE}
}

@article{kahn2017uncertainty,
  title={Uncertainty-aware reinforcement learning for collision avoidance},
  author={Kahn, Gregory and Villaflor, Adam and Pong, Vitchyr and Abbeel, Pieter and Levine, Sergey},
  journal={arXiv preprint arXiv:1702.01182},
  year={2017}
}

@inproceedings{long2018towards,
  title={Towards optimally decentralized multi-robot collision avoidance via deep reinforcement learning},
  author={Long, Pinxin and Fanl, Tingxiang and Liao, Xinyi and Liu, Wenxi and Zhang, Hao and Pan, Jia},
  booktitle=icra,
  pages={6252--6259},
  year={2018},
  organization={IEEE}
}

@article{ong2016markov,
  author = {Hao Yi Ong and Mykel J. Kochenderfer},
title = {Markov decision process-based distributed conflict resolution for drone air traffic management},
journal = jgcd,
year = {2017},
volume = {40},
number = {1},
pages = {69-80},
doi = {10.2514/1.G001822},
}

@online{golding2018meterics,
  author = {Richard Golding},
  title = {Metrics to characterize
dense airspace traffic},
  year = 2018,
  url = {http://bit.ly/altiscopetr004},
  urldate = {2019-02-03}
}

@inproceedings{schulman2015trust,
  title={Trust region policy optimization},
  author={Schulman, John and Levine, Sergey and Abbeel, Pieter and Jordan, Michael and Moritz, Philipp},
  booktitle=icml,
  pages={1889--1897},
  year={2015}
}

@article{schulman2017proximal,
  title={Proximal policy optimization algorithms},
  author={Schulman, John and Wolski, Filip and Dhariwal, Prafulla and Radford, Alec and Klimov, Oleg},
  journal={arXiv preprint arXiv:1707.06347},
  year={2017}
}

@article{seok2017task,
  title={Task selection for radar resource management in dynamic environments},
  author={Seok, Jinwoo and Kabamba, Pierre and Girard, Anouck},
  journal={The Journal of Engineering},
  volume={2018},
  number={1},
  pages={1--9},
  year={2017},
  publisher={IET}
}

@inproceedings{radanovic2017self,
  title={Self-reorganized supporting tools for conflict resolution in high-density airspace volumes},
  author={Radanovic, Marko and Eroles, MA Piera and Koca, Thimjo and Nieto, FJ Saez},
  booktitle={Twelfth USA/Europe Air Traffic Management Research and Development Seminar},
  pages={10},
  year={2017}
}

\section*{Author Biographies}
\vspace*{-1.5cm}
\begin{IEEEbiography}[{\includegraphics[width=1in,height=1in,clip]{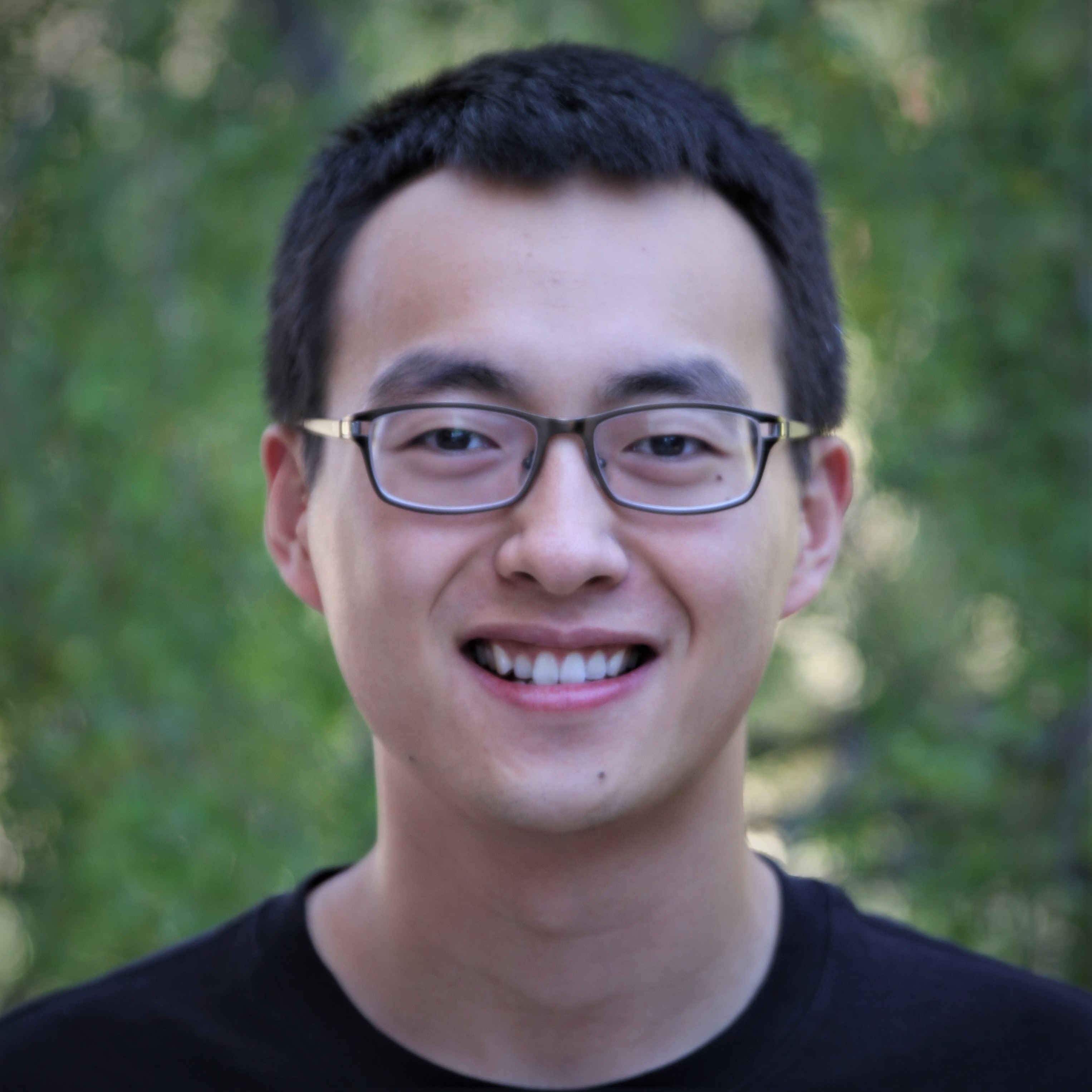}}]{Sheng Li}
is a graduate student in Aeronautics and Astronautics at Stanford University. He received his B.S.Eng. in aerospace engineering from the University of Michigan, and his B.S.Eng in mechanical engineering from Shanghai Jiao Tong University in 2017. He is currently pursuing his M.S. degree at Stanford. 
\end{IEEEbiography}
\vspace*{-1.5cm}
\begin{IEEEbiography}[{\includegraphics[width=1in,height=1in,clip]{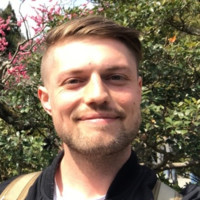}}]{Maxim Egorov} is a senior engineer at Airbus UTM where he works on decision making system for unmanned aviation. Prior to joining Airbus, he was a graduate student at Stanford University where he received his M.S. in Aeronautics and Astronautics in 2017. He received his B.S. in physics from UC Berkeley in 2013. 
\end{IEEEbiography}
\vspace*{-1.5cm}
\begin{IEEEbiography}[{\includegraphics[width=1in,height=1.1in,clip]{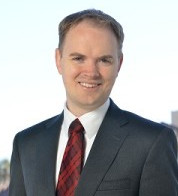}}]{Mykel Kochenderfer} is Assistant Professor of Aeronautics and Astronautics at Stanford University. Prior to joining the faculty in 2013, he was at MIT Lincoln Laboratory where he worked on airspace modeling and aircraft collision avoidance. He received his Ph.D. from the University of Edinburgh in 2006. He received B.S. and M.S. degrees in computer science from Stanford University in 2003.
\end{IEEEbiography}

\end{document}